\documentclass[twoside]{article} 

\usepackage[accepted]{aistats2017}
\usepackage{color}
\usepackage{graphicx}

\usepackage{amssymb,amsmath}
\usepackage{juremath}
\usepackage{IEEEtrantools}

\usepackage{amsthm}
\usepackage{mathtools}
\newtheorem{theorem}{Theorem}

\newtheorem{definition}{Definition}
\newtheorem{remark}{Remark}

\usepackage{booktabs}       
\usepackage{subfigure}
\usepackage{tikz}

\usepackage[sort]{natbib}

\begin{document}

%

%
\runningauthor{Jure Sokoli{\' c}, Raja Giryes, Guillermo Sapiro, Miguel R. D. Rodrigues}

\twocolumn[

\aistatstitle{Generalization Error of Invariant Classifiers}



\aistatsauthor{\centering Jure Sokoli{\' c}$^1$ \qquad Raja Giryes$^2$ \qquad  Guillermo Sapiro$^3$ \qquad  Miguel R. D. Rodrigues$^1$}
\aistatsaddress{$^1$University College London \And $^2$Tel-Aviv University \And $^3$Duke University} 



]

\begin{abstract}
This paper studies the generalization error of invariant classifiers. In particular, we consider the common scenario where  the classification task is invariant to certain transformations of the input, and that the classifier is constructed (or learned) to be invariant to these transformations. Our approach relies on factoring the input space into a product of a base space and a set of transformations. We show that whereas the generalization error of a non-invariant classifier is proportional to the complexity of the input space, the generalization error of an invariant classifier is proportional to the complexity of the base space. We also derive a set of sufficient conditions on the geometry of the base space and the set of transformations that ensure that the complexity of the base space is much smaller than the complexity of the input space. Our analysis applies to general classifiers such as convolutional neural networks. We demonstrate the implications of the developed theory for such classifiers with experiments on the MNIST and CIFAR-10 datasets.
\end{abstract}


\section{Introduction}

One of the fundamental topics in statistical learning theory is 
the one of the \textit{generalization error} (GE). Given a training set and a hypothesis class, a learning algorithm chooses a hypothesis based on the training set in such a way that it minimizes an empirical loss. This loss, which is calculated on the training set, is also called the training loss and it often underestimates  the expected loss. The GE is the difference between the empirical loss and the expected loss.

There are various approaches in the literature that aim at bounding the GE via the complexity measures of the hypothesis class, such as the VC-dimenension \citep{Vapnik1991,Vapnik1999}, the fat-shattering dimension \citep{Alon1997}, and the Rademacher and the Gaussian complexities \citep{Bartlett2003}. Another line of work provides the GE bounds based on the stability of the algorithms, by measuring how sensitive is the output to the removal or change of a single training sample \citep{Bousquet2002}. Finally, there is a recent work by \cite{Xu2012a} that bounds the GE in terms of the notion of algorithmic robustness.

An important property of the (traditional) GE bounds is that they are distribution agnostic, i.e., they hold for any distribution on the sample space. Moreover, GE bounds can lead to a principled derivation of learning algorithms with GE guarantees, e.g., Support Vector Machine (SVM) \citep{Cortes1995} and its extension to non-linear classification with kernel machines \citep{Hofmann2008}.

However, the design of learning algorithms in practice does not rely only on the complexity measures of the hypothesis class, but it also relies on exploiting the underlying structure present in the data. A prominent example is associated with the field of computer vision where the features and learning algorithms are designed to be invariant to the intrinsic variability in the data \citep{Soatto2016}. Image classification is a particular computer vision task that requires representations that are invariant to various nuisances/transformations such as viewpoint and illumination variations commonly present in the set of natural images, but do not contain ``helpful information'' as to the identity of the classified object. This motivates us to develop a theory for learning algorithms that are invariant to certain sets of transformations.

The GE of invariant methods has been studied via the VC-dimension by \cite{Abu-Mostafa1993}, where it is shown that the subset of an hypothesis class that is invariant to certain transformations is smaller than the general hypothesis class. Therefore, it has a smaller VC-dimension. Yet, the authors do not provide any characterization of  how much smaller the VC-dimension of an invariant method might be. Similarly, group symmetry in data distribution was also explored in the problem of covariance estimation, where it is shown that leveraging group symmetry leads to gains in sample complexity of the covariance matrix estimation \citep{Soloveychik2014,Shah2012a}.

There are various other examples in the literature that aims to understand/leverage the role of invariance in data processing. For example, Convolutional Neural Networks (CNNs) -- which are known to achieve state of the art results in image recognition, speech recognition, and many other tasks \citep{LeCun2015} -- are known to possess certain invariances. The invariance in CNNs is achieved by careful design of the architecture so that it is (approximately) invariant to various transformations such as  rotation, scale and affine deformations \citep{Gens2014,Cohen2016,Dieleman2016}; or by training with augmented training set, meaning the training set is augmented with some transformed versions of the training samples,  so that the learned network is approximately invariant \citep{Simard2003a}. Another example of a translation invariant method is the scattering transform, which is a CNN-like transform based on wavelets and point-wise non-linearities \citep{Bruna2012}. See also \citep{Sifre2013,Wiatowski2015a}. In practice, such learning techniques achieve a lower GE than their ``non-invariant'' counterparts.

\cite{Poggio2012a} and \cite{Anselmi2014a,Anselmi2016}  study biologically plausible learning of invariant representations and connect their results to CNNs. The role of convolutions and pooling in the context of natural images is also studied by \cite{Cohen2016a}.

There are various works that that study the GE of CNNs \citep{Sokolic2016,Shalev-Shwartz2014,Neyshabur2015,Huang2015}, however, they do not establish any connection between the network's invariance and its GE. 

Motivated by the above examples, this work proposes a theoretical framework to study the GE of invariant learning algorithms and shows that an invariant learning technique may have a much smaller GE than a non-invariant learning technique. Moreover, our work directly relates the difference in GE bounds to the size of the set of transformations that a learning algorithm is invariant to. Our approach is significantly different from  \citep{Abu-Mostafa1993} because it focuses on the complexity of the data, rather than on the complexity of the hypothesis class.

\subsection{Contributions}

The main contribution of this paper can be summarized as follows:

\vspace{0.cm}
\textit{
We prove that given a learning method invariant to a set of transformations of size $T$, the GE of this method may be up to a factor $\sqrt{T}$ smaller than the GE of a non-invariant learning method.
}
\vspace{0.cm}

Additionally, our other contributions include:
\begin{itemize}
	\item We define notions of stable invariant classifiers and provide GE bounds for such classifiers;
	\item We establish a set of sufficient conditions that ensure that the bound of the GE of a stable invariant classifier is much smaller than the GE of a robust non-invariant classifier. We are not aware of any other works in the literature that achieve this;
	\item Our theory also suggests that explicitly enforcing invariance when training the networks should improve the generalization of the learning algorithm. The theoretical results are supported by experiments on the MNIST and CIFAR-10 datasets. 
\end{itemize}
\section{Problem Statement}

We start by describing the problem of supervised learning and its associated GE. Then we define the notions of invariance in the classification task and the notion of an invariant algorithm.

\subsection{Generalization Error}

We consider learning a classifier from training samples. In particular, we assume that there is a probability distribution $P$ defined on the sample space $\sZ$ and that we have a training set drawn i.i.d. from $P$  denoted by $S_m =\{ s_i \}_{i=1}^m$, $s_i \in \sZ$, $i = 1, \ldots, m$. A learning algorithm $\sA$ takes the training set $S_m$ and maps it to a learned hypothesis $\sA_{S_m}$. The loss function of an hypothesis $\sA_{S_m}$ on the sample $z \in \sZ$ is denoted by $l(\sA_{S_m}, z)$. The  empirical loss and the expected loss of the learned hypothesis $\sA_{S_m}$ are defined as
\begin{IEEEeqnarray}{l}
	l_{\text{emp}}(\sA_{S_m}) = 1/m \sum_{s_i \in S_m} l \left(\sA_{S_m}, s_i \right) \quad \text{and} \\
	l_{\text{exp}}(\sA_{S_m}) = \E_{s \sim P} \left[ l\left( \sA_{S_m}, s \right) \right],
\end{IEEEeqnarray}
respectively; and the GE is defined as 
\begin{IEEEeqnarray}{rCl}
	GE(\sA_{S_m}) = | l_{\text{emp}}(\sA_{S_m}) - l_{\text{exp}}(\sA_{S_m}) | \,. \label{eq:GE}
\end{IEEEeqnarray}

We consider a classification problem, where the sample space $\sZ = \sX \times \sY$ is a product of the input space $\sX$ and the label space $\sY$, where a vector $\bx \in \sX \subseteq \Ro{N}$ represents an observation that has a corresponding class label $y \in \sY = \{1,2,\ldots, N_{\sY}\}$. We will write $z = (\bx, y)$ and $s_i = (\bx_i, y_i)$.

\subsection{Stable Classifier and its Generalization}

The feature extractor (e.g., CNN) used in this work defines the non-linear function $f(\bx, \theta): \Ro{N} \to \Ro{N_\sY}$, where $N_\sY$ represents the number of classes, $N$ represents the dimension of the input signal, and $\theta$ represents the parameters of the feature extractor. The classifier defined by the feature extractor is then given as 
\begin{IEEEeqnarray}{rCl}
	\argmax_{i \in [N_\sY]} (f(\bx, \theta))_i \,,
\end{IEEEeqnarray}
where, $(f(\bx, \theta))_i$ is the $i$-th element of $f(\bx, \theta)$. For example, this may correspond to a CNN with a softmax layer at the end. We will often write $f(\bx, \theta) = f(\bx)$, and define its Jacobian matrix as
\begin{IEEEeqnarray}{rCl}
	\bJ(\bx, \theta) = \frac{d f(\bx,\theta)}{d \bx} = \bJ(\bx) \,. \label{eq:jacobian_matrix}
\end{IEEEeqnarray}

A learning algorithm $\sA$ therefore returns a hypothesis, which is a function of the training set $S_m$,
\begin{IEEEeqnarray}{rCl}
	\sA_{S_m}(\bx) = \argmax_{i \in [N_\sY]} (f(\bx, \theta(S_m)))_i \,. \label{eq:classifier}
\end{IEEEeqnarray}
In a classification task, the goal of learning is to find an hypothesis that separates training samples from different classes. To model this we define the score of a training sample, which measures how confident the prediction of a classifier is:
\begin{definition}[Score] 
Consider a training sample $s_i = (\bx_i, y_i)$. The score of training sample $s_i$ is defined as
\begin{IEEEeqnarray}{rCl}
o(s_i) &=& \min_{j \neq y_i} \sqrt{2} \left( (f(\bx_i))_{y_i} - (f(\bx_i))_{j} \right) \,. \label{eq:output_score}
\end{IEEEeqnarray}
\end{definition}
Note that a large score of training samples does not imply that the learned hypothesis will have a small GE. In this work we leverage the (non-invariant) GE bounds provided by \cite{Sokolic2016}. Before providing such bounds we define the notion of learning algorithm stability and the notion of covering number that are crucial for the GE bounds.

\begin{definition}[Stable learning algorithm]
Consider the algorithm $\sA$ and the hypothesis $\sA_{S_m}(\bx)$ given in \eqref{eq:classifier}. The learning algorithm $\sA$ is stable if for any training set $S_m$
\begin{IEEEeqnarray}{rCl}
	\max_{\bx \in \Rn} \| \bJ(\bx) \|_2 \leq 1  \,, \label{eq:stability_assumption}
\end{IEEEeqnarray}
where $\| \cdot| \|_2$ denotes the spectral norm.
\end{definition}
Stability of a learning algorithm defined in this way ensures that a learned classifier has a small GE as we shall see in Theorem~\ref{th:GE_noninvariant}.

We also need a measure of complexity/size of the input space $\sX$, which is given by the covering number.
\begin{definition}
Consider a space $\sX$ and a metric $d$. We say that the set $\sC$ is an $\epsilon$-cover of $\sX$ if $\forall \bx \in \sX$, $\exists \bx' \in \sC$ such that $d(\bx, \bx') \leq \epsilon$. The covering number of $\sX$ corresponds to the cardinality of the smallest $\sC$ that covers $\sX$. It is denoted by $\N(\sX; d,\epsilon)$.
\end{definition}
In this work we will assume that $d$ is the Euclidean metric: $	d(\bx, \bx') = \| \bx - \bx' \|_2$.

Finally, we can provide the GE bounds for the stable learning algorithm. This is a variation of theorems 2 and 4 by \cite{Sokolic2016}.\footnote{\cite{Sokolic2016} also provide tighter GE bounds. For the sake of simplicity we use the bounds based on the spectral norm of the Jacobian matrix.}
\begin{theorem} \label{th:GE_noninvariant}
	Assume that the learning algorithm $\sA$ is stable and that there exists a constant $\gamma$ such that
	\begin{IEEEeqnarray}{rCl}
		o(s_i) \geq \gamma \quad \forall s_i \in S_m \,.
	\end{IEEEeqnarray}
Assume also that the loss $l(\cdot)$ is the 0-1 loss. Then, for any $\delta > 0$, with probability at least $1- \delta$, 
	\begin{IEEEeqnarray}{rCl}
	GE(\sA_{S_m}) &\leq& \sqrt{\frac{2 \log(2) \cdot N_\sY \cdot \N(\sX; d, \gamma / 2 ) }{ m}} \nonumber \\ && + \sqrt{ \frac{2 \log(1/\delta)}{m}}   \,.  \label{eq:GE_bound}
	\end{IEEEeqnarray}
\begin{proof}
The proof is straightforward by the application of theorems 2 and 4 by \cite{Sokolic2016}.
\end{proof}
\end{theorem}
The GE therefore approaches zero with rate $1/\sqrt{m}$ and it depends on the number of classes via $\sqrt{N_\sY}$. Critical is  the dependence on the covering number $\N(\sX; d, \gamma / 2 )$, which is a function of the input space $\sX$ and the margin $\gamma$. 

\subsection{Structured Input Space and Invariant Algorithms}

The bound of the GE provided in the previous section depends on the covering of the input space $\sX$. As noted in the introduction, $\sX$ often exhibits symmetries that may reduce its ``effective'' complexity and therefore also reduce the GE. We formalize this intuition in this section.

\begin{figure}[t]
\vspace{-0.1cm}
\centering
	\subfigure[Input space.]{
	\includegraphics[height=0.5\linewidth]{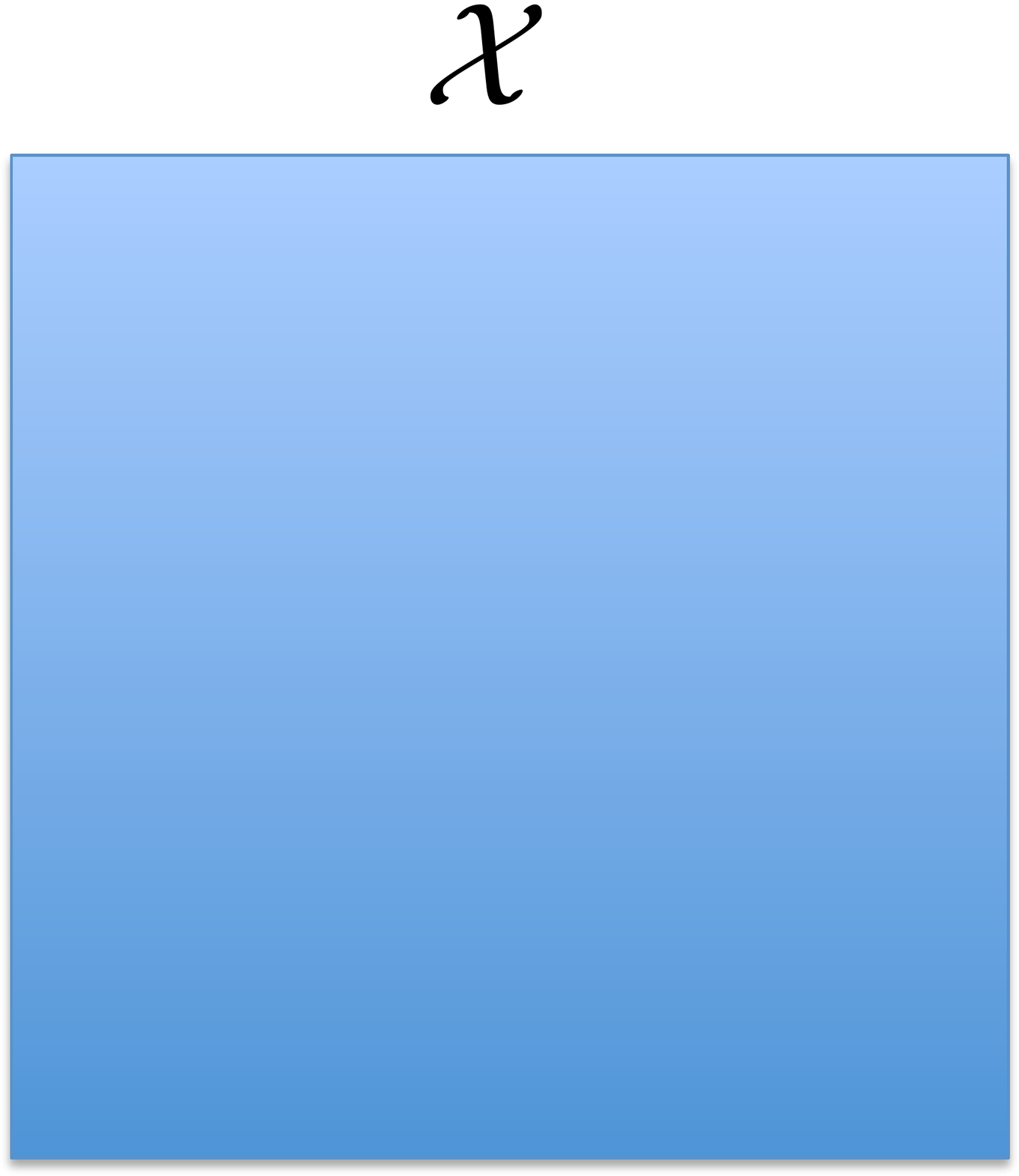}}
	\subfigure[Input space decomposition.]{
	\includegraphics[height=0.5\linewidth]{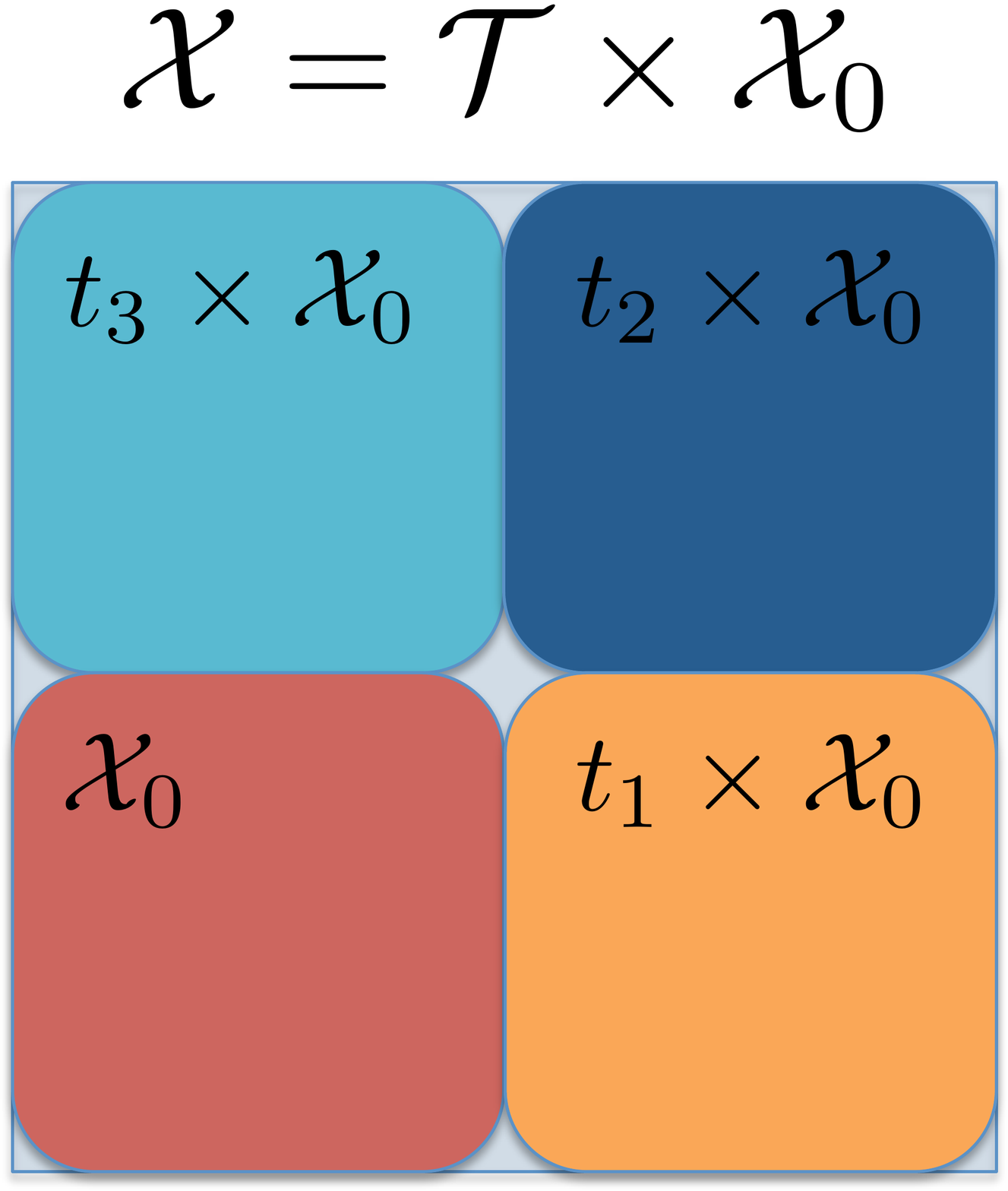}}	
	\caption{Theorem 1 shows that the size of the input space $\sX$ determines the GE of a stable learning algorithm. The input space can often be constructed as a product of a simpler base space $\sX_0$ and a set of transformations $\sT$, where the transformations in $\sT$ preserve the class labels. Theorem 2 shows that the GE of an invariant stable learning algorithm is determined by the size of the base space $\sX_0$. The size of the base space $\sX_0$ can be much smaller than the size of the input space $\sX$.} \label{fig:set_decomposition}
\vspace{-0.2cm}
\end{figure}

To capture the additional structure present in the data, we model the input space $\sX$ as a product of a base space $\sX_0$ and a set of transformations $\sT$:
\begin{IEEEeqnarray}{rCl}
	\sX = \sT \times \sX_0  \coloneqq \{t(\bx): t \in \sT, \bx \in \sX_0 \} \,, \label{eq:Xdecomposition}
\end{IEEEeqnarray}
where $\sX_0 \subseteq \Rn$, $\sT = \{ t_0, t_2, \ldots t_{T-1} \}$ and $T$ corresponds to the size of $\sT$.\footnote{Note that the discrete representation of this set is not limiting in practice.} We assume $t_0$ to be the identity, i.e., $t_0(\bx) = \bx$ throughout this work. For example, if $\sX_0$ is a set of images and $\sT$ is a set of translations, then $\sX$ will be the set of images with all possible translations of the images in $\sX_0$. See also Figure~\ref{fig:set_decomposition}.

We assume that  the classification task is invariant to the set of transformations $\sT$, i.e., we are really interested only in the set $\sX_0$ but have access to transformed samples of it, where clearly all of them have the same label. In other words, the class labels of $t(\bx)$ are the same for all $t \in \sT$. In this case, it is reasonable to leverage this by using an invariant learning algorithm.\footnote{Here we define a notion of absolute invariance. It is easy to extend it to approximate invariance, where in $\sX$ we have transformed versions of $\sX_0$ plus small/bounded noise; and  also to extend the GE bounds in a similar manner for approximately invariant learning algorithms.} 
\begin{definition}[Invariant algorithm]
A learning algorithm $\sA$ is invariant to the set of transformations $\sT$ if the embedding is invariant:
\begin{IEEEeqnarray}{rCl}
	f(t_i(\bx), S_m) = f(t_j(\bx), S_m) \quad  \forall \bx \in \sX_0,  t_i, t_j \in \sT \,, \nonumber \\ \*
\end{IEEEeqnarray}
for any training set $S_m$. We will denote such learning algorithm  by $\sA_{S_m}^\sT$.
\end{definition}
This leads us to the question that will occupy us throughout this paper:  what is the GE of an invariant learning algorithm. 

\section{Generalization Error of Invariant Classifiers}

In this section we provide bounds to the GE of invariant algorithms. The invariance of the learning method induces a possibly more efficient covering of the input space $\sX$, which translates into a lower GE.

The GE of invariant and stable learning algorithms can be bounded as follows:

\begin{theorem} \label{th:GE_invariant}
	Assume that the learning algorithm $\sA$ is stable and invariant to $\sT$ and that there exists a constant $\gamma$ such that
	\begin{IEEEeqnarray}{rCl}
		o(s_i) \geq \gamma \quad \forall s_i \in S_m \,.
	\end{IEEEeqnarray}
Assume also that the loss $l(\cdot)$ is the 0-1 loss. Then, for any $\delta > 0$, with probability at least $1- \delta$, 
	\begin{IEEEeqnarray}{rCl}
		GE(\sA_{S_m}^\sT) &\leq& \sqrt{\frac{2 \log(2) \cdot N_\sY \cdot \N(\sX_0; d, \gamma / 2 ) }{ m}} \nonumber \\ && + \sqrt{ \frac{2 \log(1/\delta)}{m}}   \,.  \label{eq:GE_bound_inv}
	\end{IEEEeqnarray}
\begin{proof}
We show that under the assumptions of this theorem the learning algorithms is $(\N(\sX_0; d, \gamma/2), 0)$-robust (see \citep{Xu2012a} or \citep{Sokolic2016}). The GE bound then follows from Theorem 3 and Example 9 by \cite{Xu2012a} (or theorems 1 and 2 by \cite{Sokolic2016}).

We construct a covering as follows. Take the covering that leads to the covering number $\N(\sX_0; d, \gamma/2)$ and denote the subsets of $\sX_0$ by $\sK_i$, $i = 1, \ldots, \N(\sX_0; d, \gamma/2)$. By the definition of $\sX$ in \eqref{eq:Xdecomposition} we can cover $\sX$ by $\N(\sX_0; d, \gamma/2)$ sets of the form $\sT \times \sK_i$, $i = 1, \ldots, \N(\sX_0; d, \gamma/2)$.

Now take $\bx_i$ in the training set and $\bx \in \sX$ such that $\bx_i, \bx \in \sT \times \sK_j$. Due to the invariance of $f$ we have $\| f(\bx_i) - f(\bx) \|_2 < \gamma$ and all $\bx$ will lie in the same decision region as $\bx_i$. This implies that stable and invariant learning algorithm is  $(\N(\sX_0; d, \gamma/2), 0)$-robust. The GE bound follows from Theorem 3 by \cite{Xu2012a}.
\end{proof}
\end{theorem}

Note that the GE bound in Theorem~\ref{th:GE_invariant} is of the same form as the GE bound in Theorem~\ref{th:GE_noninvariant} and the main difference is in the employed covering number. In particular, the ratio between the bounds is
\begin{IEEEeqnarray}{rCl}
	R(\sX_0, \sX; d, \epsilon) = \left( \frac{ \N(\sX_0; d, \epsilon )}{ \N(\sX; d, \epsilon )} \right)^{1/2} \,,
\end{IEEEeqnarray}
where $\epsilon =  \gamma /2$ in our case. We are especially interested in the scenarios where the GE bound of an invariant method is much smaller than the GE bound of a non-invariant method. This happens when \mbox{$R(\sX_0, \sX; d , \epsilon) \ll 1$}. We now establish a set of sufficient conditions on $\sX_0$, $\sT$, $d$ and $\epsilon$ such that \mbox{$R(\sX_0, \sX; d, \epsilon) \ll 1$}.

\begin{theorem} \label{th:ratio_bound}
	Assume that $\sX = \sT \times \sX_0$ and choose $\epsilon < 1$. Then
\begin{IEEEeqnarray}{c}
d(t(\bx), t'(\bx')) > 2 \epsilon \quad \forall \bx, \bx' \in \sX_0, t \neq t' \in \sT  \label{eq:covering_condition} \\
	\text{and} \nonumber \\
d(t(\bx), t(\bx')) \geq d(\bx, \bx') \quad \forall \bx, \bx' \in \sX_0, t  \in \sT \label{eq:isometry_condition} \\
\implies 
R(\sX_0, \sX; d, \epsilon) \leq 1/ \sqrt{T} \,, \label{eq:ratio_bound}
	\end{IEEEeqnarray}	
	
where $T$ is the number of elements in $\sT$. On the other hand, 
\begin{IEEEeqnarray}{C}
d(t(\bx), t'(\bx)) = 0 \quad \forall \bx \in \sX_0, t \neq t' \in \sT \label{eq:same_condition}  \\
\implies  R(\sX_0, \sX; d, \epsilon) = 1. \label{eq:ratio_bound_2}
\end{IEEEeqnarray}
\begin{proof}
Consider any covering of $\sX_0$ that leads to the covering number $\N(\sX_0; d, \epsilon )$. Denote the metric balls of radius $\epsilon$ that cover $\sX_0$ by $\sC_i$, $i = 1, \ldots, \N(\sX_0; d, \epsilon )$.  Denote the elements of $\sT$ as $t_j$, $j = 1, \ldots, T$ and the transformed sets by $t_j(\sX_0) = \{t_j(\bx) : \bx \in \sX_0 \}$, $j = 1, \ldots, T$.

First, we show that \eqref{eq:covering_condition} implies that any possible metric ball of radius $\epsilon$ can only have non-empty intersection with one of the ``copies'' of $\sX_0$. Denote by $\sB$ an arbitrary metric ball of radius $\epsilon$. Then
\begin{IEEEeqnarray}{rCl}
	\sB \cap t_j(\sX_0) \not= \emptyset \implies 	\sB \cap t_k(\sX_0) = \emptyset \quad \forall k \neq j.
\end{IEEEeqnarray}
To see this, observe that the definition of $\sB$ implies  that $d(\bx, \bx') \leq 2 \epsilon$, $\forall \bx, \bx' \in \sB$. Now take a point $\bx \in \sB \cap t_j(\sX_0)$ and a point $\bx' \in t_k(\sX_0)$, $k \neq j$. Note that by \eqref{eq:covering_condition} $d(\bx, \bx') > 2 \epsilon$, which implies that $\bx' \not \in \sB $ and therefore $\sB \cap t_k(\sX_0) = \emptyset$. This implies that the  covering number of $\sX$ with metric ball of radius $\epsilon$ is
\begin{IEEEeqnarray}{rCl}
	\N(\sX; d, \epsilon) = \sum_{j=1}^T \N(t_j(\sX_0); d, \epsilon) \,.
\end{IEEEeqnarray}
Finally, it remains to be proven that $\N(t_j(\sX_0); d, \epsilon) \geq \N(\sX_0; d, \epsilon)$ $\forall t_j \in \sT$, which is straightforward to establish given the condition  \eqref{eq:isometry_condition}. This proves \eqref{eq:ratio_bound}. Proof of \eqref{eq:ratio_bound_2} is trivial as $\sX_0 =  \sX$ when \eqref{eq:same_condition} holds. 
\end{proof}
\end{theorem}

We have shown, via conditions on the geometry of the base space $\sX_0$, and the effect of transformations in $\sT$ on it, that the ratio $R(\sX_0, \sX; d, \epsilon)$ can be smaller or equal to $1/\sqrt{T}$. Note that conditions \eqref{eq:covering_condition} and \eqref{eq:isometry_condition} ensure that the effect of transformations in $\sT$ can not be captured by the metric $d$. Otherwise, the invariant algorithm has no advantage over a non-invariant one (this is illustrated by examples in Section~\ref{sec:examples}):
\begin{itemize}
	\item The sufficient condition in  \eqref{eq:covering_condition} can be stated as follows. Take any pair of vectors in the base space $\sX_0$ and transform them by the two transformations in $\sT$ that are not equal. Then the distance between the pair of vectors must be at least $2 \epsilon$. In other words, the transformation must not make two distinct vectors in the base space $\sX_0$ (distance at least $2 \epsilon$) indistinguishable (distance smaller than $2 \epsilon$). Similarly, any two transformations in $\sT$ that are not equal must make two similar vectors in $\sX_0$ (with distance smaller than $2 \epsilon$) distinct  (distance at least $2 \epsilon$). 
	
	\item The sufficient condition in  \eqref{eq:isometry_condition} ensures that the transformations in $\sT$ are not trivial, i.e., they do not reduce the complexity of the base space $\sX_0$. For example, a transformation that maps any $\bx \in \sX_0$ into itself violates  \eqref{eq:isometry_condition} and leads to a set of the same complexity, as formalized by \eqref{eq:same_condition}.
\end{itemize}

The results of this section can be summarized by the following remark:
\begin{remark}
Given an input space  $\sX$, which is structured according to the assumptions of Theorem~\ref{th:ratio_bound} and the size of transformation set $T$, we have established that the GE of an invariant stable learning algorithm may be up to a factor $\sqrt{T}$ smaller than the GE of a non-invariant stable learning algorithm. To the best of our knowledge, this is the first time such quantitative result is provided for invariant algorithms. 

\end{remark}

\subsection{Illustration} \label{sec:examples}
\begin{figure}[t]
	\centering
	\subfigure[Atoms.]{
	\includegraphics[height=0.4\linewidth]{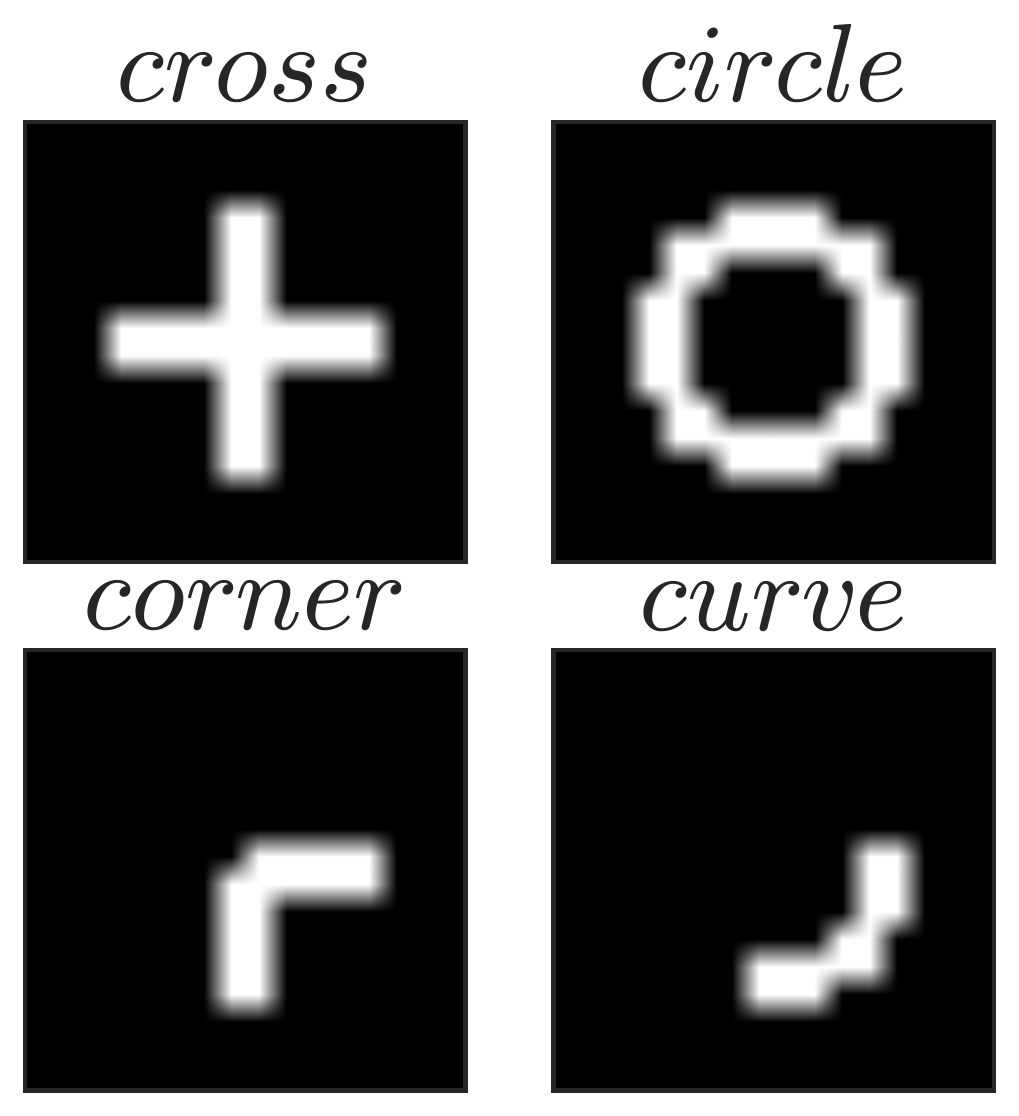}}
	\hspace{1cm}
	\subfigure[Transformed atoms.]{
	\includegraphics[height=0.4\linewidth]{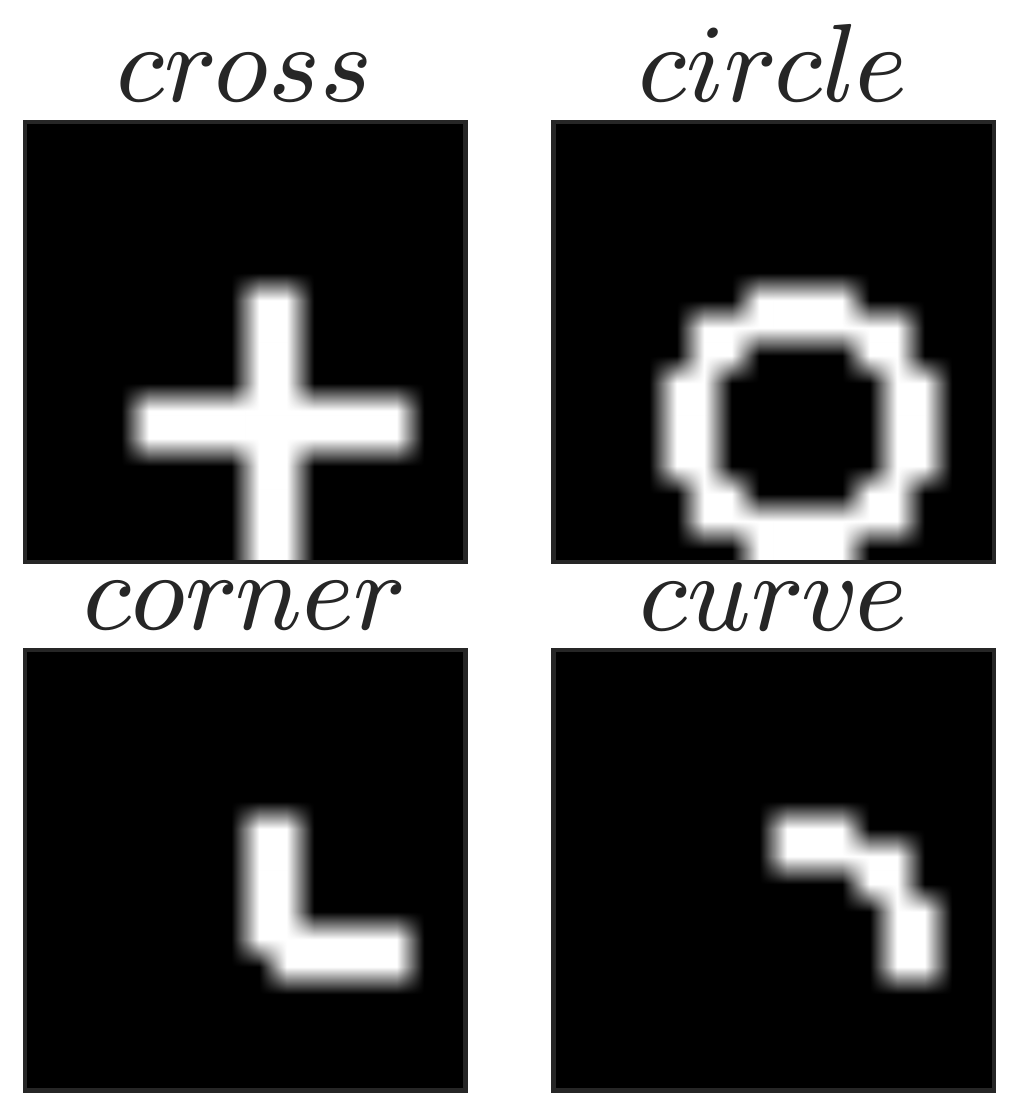}}	
	\caption{(a) A set of atoms (\textit{cross}, \textit{circle}, \textit{corner}, \textit{curve}) used to construct the base space. (b) Examples of transformed atoms with a transformation from the trans-rotation set.} \label{fig:atoms}
\vspace{-0.1cm}
\end{figure}

To provide additional intuition related to Theorem 3, we present the following toy example. We consider four images of dimension $N \times N$, with $N = 16$, Figure~\ref{fig:atoms}(a). The sets of transformations that we consider are: 
\begin{itemize}
	\item Translation set: The set of pixel-wise cyclic translations in any direction. The size of the set is $N^2$. 
	
	\item Rotation set: The set of image rotations by $90^\circ$. The size of this set is 4 (this may explain why the $90^\circ$ rotation invariance is useful but not as critical as the translation invariance). 
	
	\item Trans-rotation set: A product of the translation and the rotation sets, where the rotation is applied first followed by a translation. The size of this set is $4 \times N^2$.
\end{itemize}
Note that all the transformations above can be implemented by permutation matrices which are orthonormal. This is important as it implies that all the considered sets  satisfy the condition in \eqref{eq:isometry_condition}. Examples of transformed atoms are shown in Figure~\ref{fig:atoms}(b).

We now provide an example of a base space $\sX_0$ and a transformation set $\sT$ for which $R(\sX_0, \sX; d, \epsilon) \leq 1/\sqrt{T}$; and then provide an example of a base space $\sX_0$ and a transformation set $\sT$ for which $R(\sX_0, \sX; d, \epsilon) \not \leq 1/\sqrt{T}$.

\paragraph{Example for $R(\sX_0, \sX; d, \epsilon) \leq 1/\sqrt{T}$:}

Consider $\sX_0 = \{cross, circle, corner, curve \}$ and $\sT$ to be the translation set. The set $\sX = \sT \times \sX_0$ then contains all possible translations of shapes in $\sX_0$. We have verified numerically that the condition in \eqref{eq:covering_condition} is satisfied for all $\epsilon < 0.375$. Therefore, $R(\sX_0, \sX; d, \epsilon) \leq 1/\sqrt{T}$ for $\epsilon < 0.375$, where $\sqrt{T} = N = 16$ is the dimension of the images. Therefore, a translation invariant learning method can attain a GE with a factor $N$ smaller than the GE of a non-invariant method.

Similarly, if we take $\sX_0 = \{corner, curve \}$ and $\sT$ to be the trans-rotation set,  we can establish  $R(\sX_0, \sX; d, \epsilon) \leq 1/(2 N)$ for $\epsilon < 0.26$.

\paragraph{Examples for $R(\sX_0, \sX; d, \epsilon) \not \leq 1/\sqrt{T}$:}

Now consider $\sX_0 = \{cross, circle \}$ and $\sT$ to be the rotation set. Therefore, $\sX = \sT \times \sX_0$ contains all possible $90^\circ$ rotations of \textit{circle} and \textit{cross} in Figure~\ref{fig:atoms}(a). It is clear that the \textit{circle} and \textit{cross} are already invariant to such rotation, i.e., they corresponds to exactly the same shape. Therefore, the condition in \eqref{eq:same_condition} holds and $R(\sX_0, \sX; d, \epsilon) = 1$. Clearly, in such cases, an invariant learning algorithm is not expected to have a smaller GE than a non-invariant learning algorithm.



\section{Invariant CNNs} 

In this section we discuss the implication of our theory on CNNs, which are very popular for classification. Note that this is one particular example and that our theory holds also for other possible classifiers.  We consider two ways for which invariance can be achieved for CNNs: via an appropriate construction of the CNN architecture or by training them to be invariant.

\paragraph{Invariance of the CNN Architecture-}

Given that the set of transformations is a group, averaging a function over a group leads to an invariant representation \citep{Anselmi2014a,Bruna2012,Cohen2016}. For example, in conventional CNNs the pooling operators usually average over the translation group and make the CNNs translation invariant.

\cite{Cohen2016} generalize the notion of convolution over translation group to general groups, which leads to architectures that can be invariant to arbitrary transformations that form discrete groups. A related approach involves normalization of the network input, which eliminates the effect of affine transformations of the input \citep{Jaderberg2015}.

\paragraph{Invariance of the CNN Learning-}

As an alternative to encoding the invariances in the CNN architecture we can train a CNN to become invariant.  This is particularly helpful in the cases that we do not know exactly how to characterize or impose the invariance manually on the network. Such an ``approximate invariance'' is achieved by training CNNs with data augmentation, which involves training the network with the transformed samples of the training examples. This was indicated by \cite{Lenc2015}, who showed that CNNs trained on the ImageNet \textit{implicitly} learn to be invariant to flips, scalings and rotations. 

Our theory suggests that enforcing the invariance of the CNN representation \textit{explicitly} should improve the robustness of CNNs and improve their GE. For example, we may train networks with an explicit regularization term of the form
\begin{IEEEeqnarray}{rCl}
	\sum_{t \in \sT} \| f(\bx_i) - f(t(\bx_i)) \|_2^2 \,, \label{eq:invariance_regularizer}
\end{IEEEeqnarray}
which promotes the invariance of the representation. We validate the effectiveness of this regularization in Section~\ref{sec:experiments_cifar}.

\section{Experiments} \label{sec:experiments}

We now demonstrate the theoretical results with experiments on the MNIST and CIFAR-10 datasets. 

\subsection{Rotation Invariant CNN}

Here we compare a rotation invariant CNN and a conventional CNN on  rotated MNIST datasetses. The rotated MNIST-$D^\circ$ dataset is constructed by rotating the digits by an angle $r \cdot D^\circ$, $r \in \{0,1,2,\ldots, 360/D - 1\}$, where the index $r$ is chosen randomly for each image in the dataset. We use $D = 180, 90, 45$.

We use a 7 layer CNN architecture: $(32,5,5)$-conv, $(2,2)$-max-pool, $(64,5,5)$-conv, $(2,2)$-max-pool, $(128,5,5)$-conv followed by a global average pooling layer and a softmax layer, where $(k,u,v)$-conv denotes the convolutional layer with $k$ filters of size $u \times v$, and $(p,p)$-max-pool denotes the max-pooling layer with pooling regions of size $p \times p$. The rotation invariant CNN is the same as the conventional CNN, but it includes a cyclic slice layer before the first convolutional layer and a cyclic pool layer before the softmax layer. Both, the cyclic slice layer and the cyclic pool layer were proposed by \cite{Dieleman2016} and together they ensure that the CNN is invariant to rotations. In particular, the cyclic slice layer takes input image $\bx$ and creates copies of $\bx$, each rotated for $r \cdot D^\circ$, $r = 0,1,2,\ldots, 360/D - 1$, where $D$ is the same as in the dataset MNIST-$D^\circ$. The copies are then passed through the CNN independently. At the end of the CNN, before the softmax layer, the outputs of the copies are averaged by a cyclic pool layer to obtain a rotation invariant representation.

The networks are trained using stochastic gradient descent (SGD) with momentum, which was set to $0.9$. The training objective is the standard categorical cross entropy (CCE) loss. Batch size was set to 32 and learning rate was set to $0.01$ and reduced by 10 after 100 epochs. The networks were trained for 150 epochs in total. Weight decay regularization was set to $10^{-4}$. We used training sets of sizes $10^3$, $10^4$, $2\cdot 10^4$, $5\cdot 10^4$. 

\begin{figure*}[t]
	\centering
	\subfigure[Test set classification acc.]{
	\includegraphics[height=0.19\paperwidth]{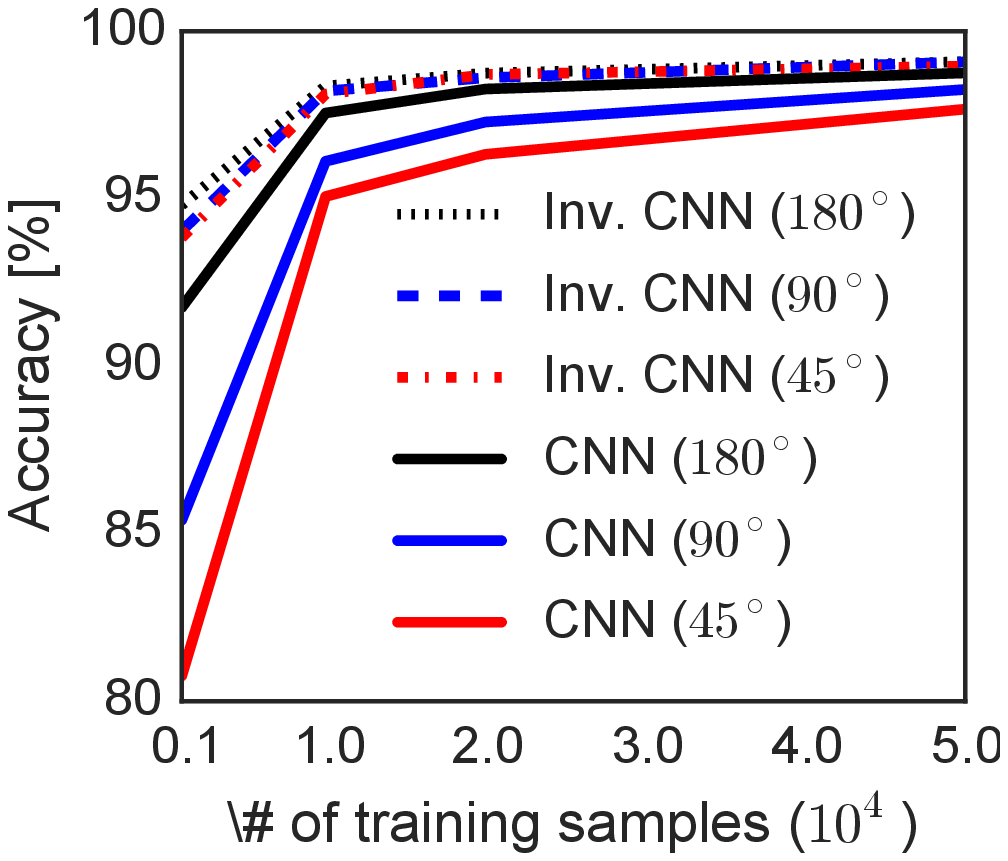}}	
	\hspace{-0.2cm}
	\subfigure[Generalization error.]{
	\includegraphics[height=0.19\paperwidth]{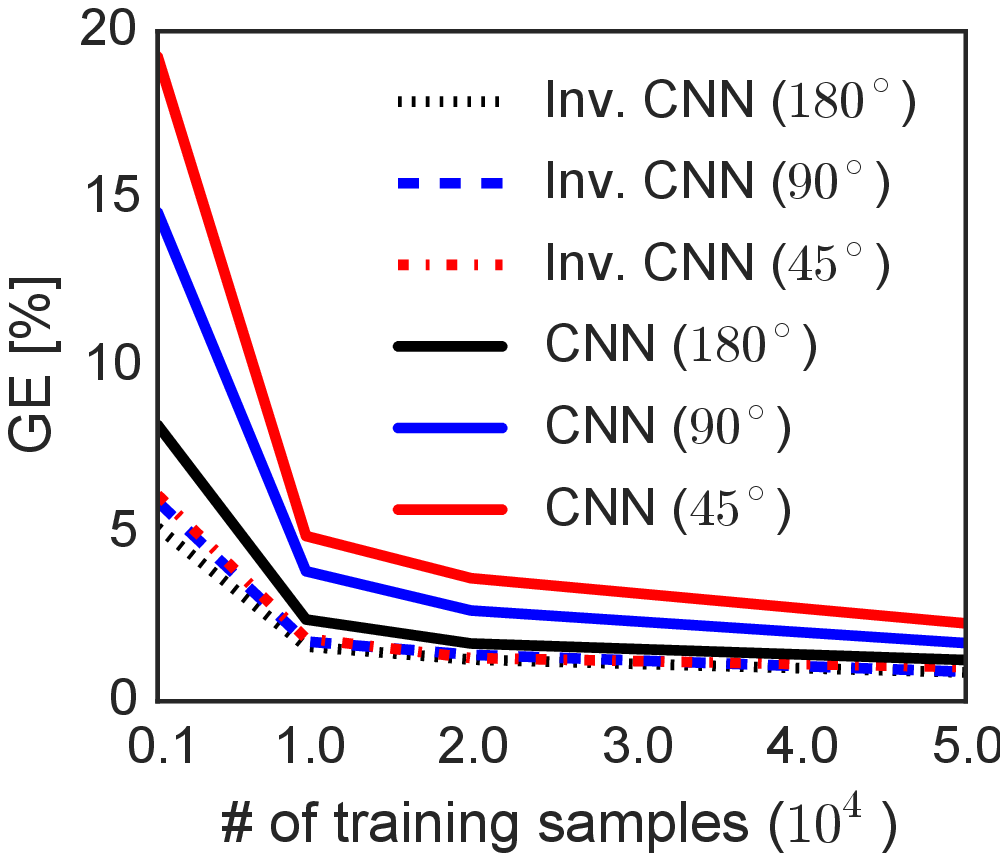}}
	\hspace{-0.2cm}	
	\subfigure[Generalization error ratio.]{
	\includegraphics[height=0.19\paperwidth]{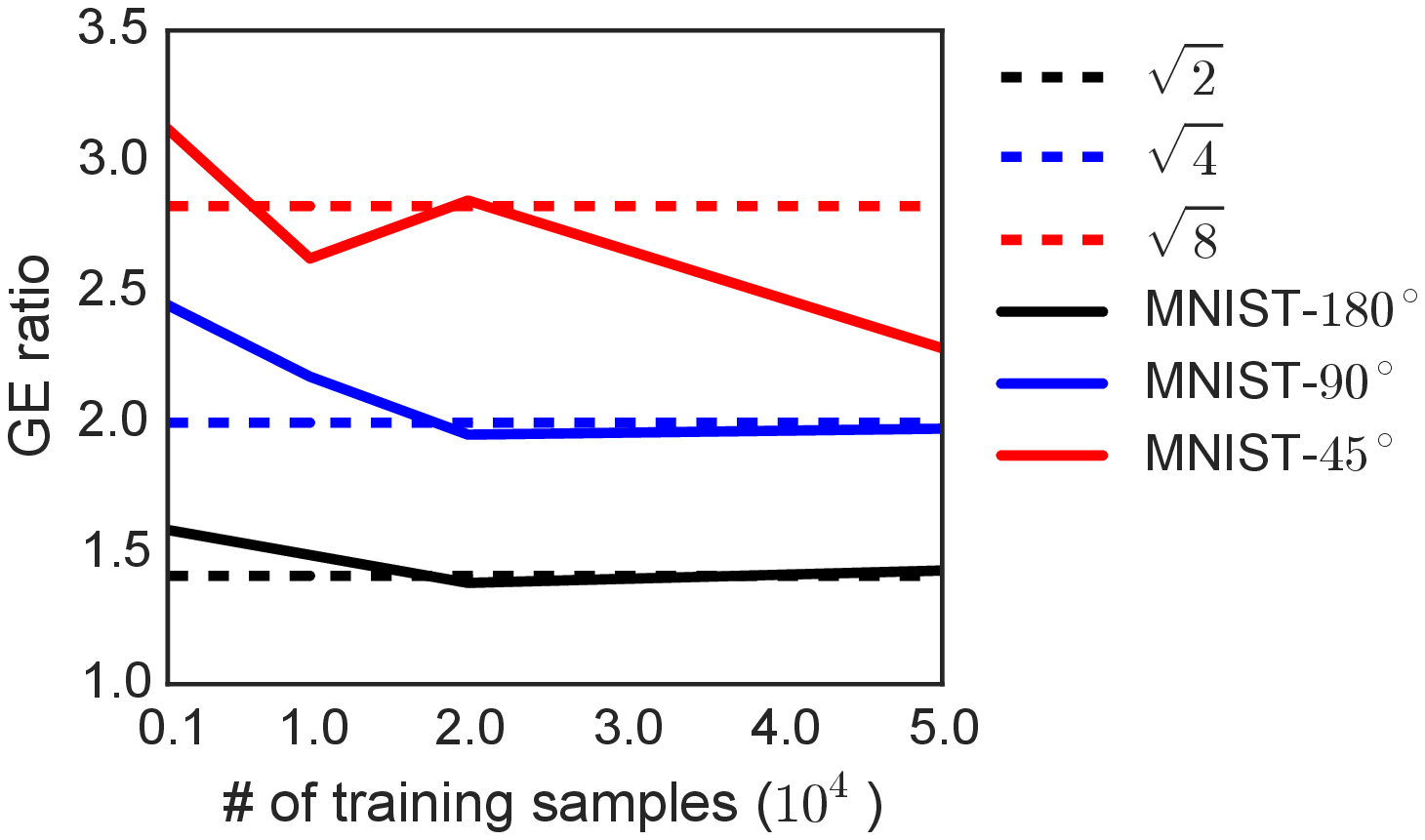}}
\vspace{-0.2cm}	
\caption{(a) Clasification accuracy, (b) the GE of the rot. invariant CNN and the conventional CNN and (c) the ratio of the GEs of the rotation invariant CNN and the conventional CNN on the rotated MNIST datasets. } \label{fig:mnist_ro}	
\vspace{-0.1cm}
\end{figure*}

The classification accuracies are reported in Figure~\ref{fig:mnist_ro}(a), the GE is reported in  Figure~\ref{fig:mnist_ro}(b) and the ratio of the GEs of the invariant and the conventional CNNs are shown in Figure~\ref{fig:mnist_ro}(c). We may note that the (explicitly) rotation invariant CNN always has a higher classification accuracy than the conventional CNN. Moreover, the GE of the rotation invariant CNN is much smaller than the GE of the conventional CNN. The difference is most significant when the training set is small, which demonstrates the importance of invariance for the generalization of learning algorithms. 

Note also that the GE of the rotation invariant CNNs on different datasets MNIST-$D^\circ$, $D=180,90,45$, is roughly the same, whereas the conventional CNNs have a higher GE on the datasets with a smaller $D$. This can be explained by the fact that the rotated MNIST dataset with a smaller $D$ is more complex due to the larger number of rotations. The sizes of the transformation sets for $D=180,90,45$ are $2,4$ and 8, respectively. Theorem~\ref{th:ratio_bound} predicts that the ratio of the GEs of an invariant and a non-invariant CNNs is equal to $\sqrt{|\sT|}$. The actual ratios are shown in Figure~\ref{fig:mnist_ro}(c). We can observe that the GE ratios obtained empirically roughly follow the theoretical prediction. However, when the training set is small, the conventional CNN generalizes worse than predicted by our theory and when the training set is large, the conventional CNN generalizes better than predicted by our theory. We conjecture that the conventional CNNs learn to be ``partially'' invariant when the number of training samples is large. Moreover, the current theory might not capture the relationship between invariant and non-invariant CNNs entirely, especially when the assumptions of Theorem 3 do not hold.

Finally, we also consider the rotation invariant MNIST dataset, where each image $\bx$ in the dataset is rotated by $r \cdot D^\circ$, $r \in \{0,1,2,\ldots, 360/D - 1\}$ and the $360/D$ copies are averaged to obtain a sample. As our theory suggests, the rotation invariant CNNs in this case do not have a lower GE than a conventional CNN because the dataset itself is rotation invariant. In fact, given the rotation invariant MNIST dataset, the rotation invariant CNN and the conventional CNN are equivalent. This can be easily established by observing that the cyclic slicing layer produces copies of the input that are identical. We have verified empirically that the rotation invariant and the non-invariant CNNs perform the same on the rotation invariant MNIST dataset.

\subsection{Learning the Invariances} \label{sec:experiments_cifar}

Finally, we demonstrate that learning invariances explicitly can lead to a lower GE. We use the CIFAR-10  dataset, which is normalized following \citep{Zagoruyko2016}, and the Wide ResNet \citep{Zagoruyko2016} with 13 layers of width 5.

The networks are trained using SGD and the learning rate is set to 0.01 for the first epoch and then to $0.05$, $0.005$ and $0.0005$, each for 30 epochs. We use $10^3$, $10^4$, $2\cdot 10^4$ and $5\cdot 10^4$ training samples. We have found that using the Jacobian regularization \citep{Sokolic2016} improves performance in all cases and it's factor is set to $0.1$ with smaller training sets (2500, 5000, 10000) and  $0.05$ otherwise. Batch size is set to 128.

SGD batches are constructed as follows: the first half of the batch contains images from the training set and the other half of the mini batch contains transformed versions of the images in the first half of the mini batch where the transformations are chosen at random. The set of transformations contains shifts of $\pm4$ pixels and horizontal flips, as in \citep{Zagoruyko2016}.

We promote the invariance by using the regularizer in \eqref{eq:invariance_regularizer}. We chose to regularize the output of the last global pooling layer instead of the softmax output and use the corresponding pairs from the batch to compute \eqref{eq:invariance_regularizer}. The regularization factor in all experiments is set to $10^{-4}$.

Table~\ref{tab:cifar} reports the standard test accuracy and the accuracy of the predictions averaged over the augmented test set (denoted by + avg.), which are obtained as follows: for each test image we average the softmax outputs for the original image, shifted images ($9 \times 9$ shifts), horizontally flipped image and scaled images (scaling factors are $0.8$ and $1.2$). Note that this method requires approximately 80 forward passes through a network to obtain a prediction.

Classification accuracies on the test set and on the augmented test set for CNNs trained with invariance regularization and for CNNs trained without the invariance regularization are reported in Table~\ref{tab:cifar}. The training set accuracies were \mbox{100\%} or very close to \mbox{100\%} in all cases. First, we observe that invariance regularization leads to a lower GE (a higher accuracy) in all cases. Moreover, testing with the augmented test set is even more robust and leads to a lower GE for both, the regularized and the non-regularized CNNs. Note however, that CNNs trained with explicit invariance regularization  (except when 2500 training samples are used) performs better or on par with a non-regularized network evaluated on the augmented test set, where testing with the augmented test set is approximately 80 times more expensive than conventional testing with a single image. This experiment verifies the hypothesis that enforcing the network invariance explicitly can lead to a smaller GE.

The ratio of the GEs of the CNN trained with data augmentation and the invariance regularization and the CNN trained without data augmentation are between $1.5$ and $2$. Note that the theory from Section~3 does not apply directly as (i) the CNN trained without data augmentation is already (partially) invariant to translations due to its convolutional structure with pooling \citep{Bruna2013b, Boureau2010}; (ii) the CNNs trained with data augmentation and invariance regularization are not perfectly invariant as defined in Definition~4, but only approximately invariant.

\begin{table}[t]
{
\centering
\caption{Classification accuracy $[\%]$ on CIFAR-10. } \label{tab:cifar}
\begin{tabular}{lccccc}
\toprule
& \multicolumn{5}{c}{number of training samples} \\
 & 2500 & 5000 & 10000 & 20000 & 50000   \\ 
   
\cmidrule(r){2-6}

No reg.  & 68.71 & 76.74 & 85.17 & 87.15 & 93.65   \\ 

Inv. Reg. & 69.32 & 79.08 & 86.69 & 88.14 & 94.50 \\ 

\midrule

{\tiny\begin{tabular}{@{}c@{}}No reg. \\ + avg. \end{tabular}} & 70.59 & 78.40 & 86.05 & 88.13 & 94.26 \\

{\tiny\begin{tabular}{@{}c@{}}Inv. Reg. \\ + avg. \end{tabular}} & 70.71 & 79.65 & 86.96 & 88.98 & 94.78   \\ 
\bottomrule
\end{tabular} 
}
\vspace{-0.2cm}
\end{table}

\section{Discussion and Conclusion}

We have formally demonstrated that the GE of an invariant learning algorithm can be much smaller than the GE of a non-invariant learning algorithm, provided that the input space can be  factorized into a product of a transformation set and a base space, where the covering number of the base space is much smaller than the covering number of the input space. This work offers an important foundation for the study of the GE of learning algorithms, such as CNNs and their extensions, that
leverage symmetries in the data.

Our assumption in this work is that the set of transformations $\sT$ is discrete. A more general approach would be to assume that the set of transformations $\sT$ is continuous. We conjecture that current results can be extended to such cases by an appropriate covering of $\sT$. Second, we have assumed that a learning method is perfectly invariant. The notion can be extended to approximately invariant learning methods and bounds of the same form as in \eqref{eq:GE_bound_inv} can be derived for this case.
\subsubsection*{Acknowledgements}
The work of J. Sokoli\'c and M. R. D. Rodrigues was supported in part by EPSRC under grant EP/K033166/1. The work of R. Giryes was supported in part by GIF. The work of G. Sapiro was supported in part by NSF, ONR, ARO, and NGA.

\clearpage
\subsubsection*{References}
\renewcommand{\section}[2]{}%
\bibliographystyle{abbrvnat_jure}
\bibliography{library_correct}


\end{document}